\documentclass[sn-mathphys]{sn-jnl}

\jyear{2021}%

\theoremstyle{thmstyleone}%
\theoremstyle{thmstyletwo}%

\theoremstyle{thmstylethree}%

\begin{document}

\title[Stable Parallel Training of Wasserstein Conditional GANs]{Stable Parallel Training of Wasserstein Conditional Generative Adversarial Neural Networks}


\author*[1]{\fnm{Massimiliano} \sur{Lupo Pasini}}\email{lupopasinim@ornl.gov}

\author[2]{\fnm{Junqi} \sur{Yin}}\email{yinj@ornl.gov}

\affil*[1]{\orgdiv{Computational Sciences and Engineering Division}, \orgname{Oak Ridge National Laboratory}, \orgaddress{\street{1 Bethel Valley Road}, \city{Oak Ridge}, \postcode{37831}, \state{TN}, \country{USA}}}

\affil[2]{\orgdiv{National Center for Computational Sciences Division}, \orgname{Oak Ridge National Laboratory}, \orgaddress{\street{1 Bethel Valley Road}, \city{Oak Ridge}, \postcode{37831}, \state{TN}, \country{USA}}}


\abstract{We use a stable {parallel} approach to train Wasserstein Conditional Generative Adversarial Neural Networks (W-CGANs). The {parallel} training reduces the risk of mode collapse and enhances scalability by using multiple generators that are concurrently trained, each one of them focusing on a single data label. The use of the Wasserstein metric reduces the risk of cycling by stabilizing the training of each generator. We apply the approach on the CIFAR10, CIFAR100, and ImageNet1k datasets, three standard benchmark image datasets. Performance is assessed using the inception score, the Fréchet inception distance, and image quality. {An improvement in inception score and Fréchet inception distance} is shown in comparison to previous results obtained by performing the {parallel} approach on deep convolutional conditional generative adversarial neural networks (DC-CGANs) as well as an improvement of image quality of the new images created by the GANs approach. Weak scaling is attained on both datasets using up to 2,000 NVIDIA V100 GPUs on the OLCF supercomputer Summit.}


\maketitle

{\footnotesize \noindent This manuscript has been authored in part by UT-Battelle, LLC, under contract DE-AC05-00OR22725 with the US Department of Energy (DOE). The US government retains and the publisher, by accepting the article for publication, acknowledges that the US government retains a nonexclusive, paid-up, irrevocable, worldwide license to publish or reproduce the published form of this manuscript, or allow others to do so, for US government purposes. DOE will provide public access to these results of federally sponsored research in accordance with the DOE Public Access Plan (\url{http://energy.gov/downloads/doe-public-access-plan}).
}

\section{Introduction}
\begin{figure}
\centering
\includegraphics[width=\textwidth]{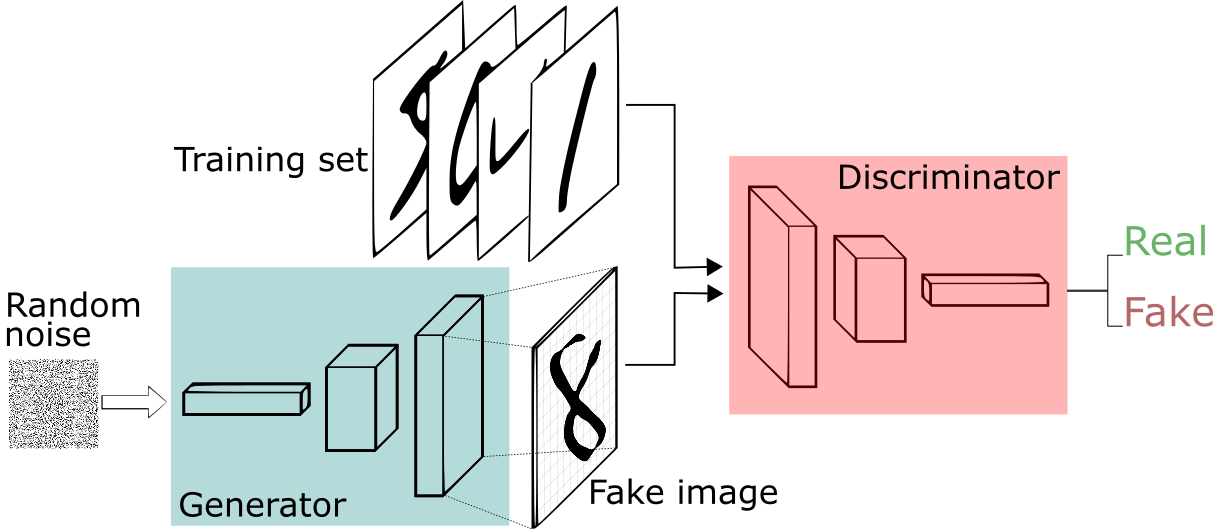}
\caption{The GAN framework pits two adversaries against each other in a game.
Each player is represented by a differentiable function controlled by a set of parameters. Typically these functions are implemented as deep neural networks. Training examples are randomly sampled from the training set and used as input for the first player, the discriminator. The goal of the discriminator is to output the probability that its input is real rather than fake, under the assumption that half of the inputs it is ever shown are real and half are fake. Image from https://sthalles.github.io/intro-to-gans/}
\label{gans_picture}
\end{figure}

Generative adversarial neural networks (GANs) \cite{goodfellow_generative_2014} \cite{radford_unsupervised_2016} \cite{salimans_improved_2016} \cite{bertsekas_multiagent_2019}  are deep learning (DL) models whereby a dataset is used by an agent, called the generator, to sample white noise from a latent space and simulate a data distribution to create new (fake) data that resemble the original data it has been trained on. Another agent, called the discriminator, has to correctly discern between the original data (provided by the external environment for training) and the fake data (produced by the generator). An illustration that describes a GANs model is shown in Figure \ref{gans_picture}. 
The generator prevails over the discriminator if, at the end of the training, the latter does not succeed in distinguishing the original from the fake. The discriminator prevails over the generator if the fake data created by the generator is categorized as fake and the original data is still categorized as original. 
The interplay between generator and discriminator can be interpreted either as a collaborative game, or a competitive game, according to the specifics of the  application. In national security applications, the generator and the discriminator are truly adversarial because the generator plays the role of a hacker that tries to breach through a security barrier and the discriminator aims at correctly distinguishing between legitimate operations of regular users against harmful illegal attacks operated by hackers. In other situations, the discriminator collaborates with the generator by helping it to improve its performance. An example is provided by data augmentation, where the generator aims at sampling from a given data distribution by extracting relevant features that can be used to produce new data samples. 
Another example occurs in healthcare, where GANs are used for the design of new drugs. In this case, the goal of the generator is to propose the composition of new drugs either to improve existing treatments or to propose treatments to types of diseases that are not curable yet, and the goal of the discriminator is to help the generator design more effective drugs by assessing their efficacy. 
GANs agents collaborate with each other also in animation applications, where the generator is in charge of creating virtual (but still realistic) representations of the reality, which are used in video games to formulate alternative scenarios, and the discriminator provides feedback to the generator to quantify how realistic the proposed scenario is. One final example of an application where the GANs training is set as a formally adversarial, but essentially a collaborative game is provided by photograph editing. In this context, GANs can be used for reconstructing images of faces to identify changes in features such as hair color, facial expressions, or gender, etc. This can help authorities identify criminals that might have undergone surgeries to modify their appearance.

In general, the training of GANs runs into two main numerical issues, namely cycling \cite{mertikopoulos_2017} and mode collapse \cite{NIPS2017_165a59f7}. Cycling happens when the generator alternates between different regions of the data space without necessarily improving its performance. Mode collapse happens when the generator gets stuck in a small region of the data space and produces the same image over time.
Avoiding cycling prevents the generator from wasting computational power by exploring the same region. Avoiding mode collapse allows the generator to escape local minima and more thoroughly explore the data space to ensure that the entire data distribution is equally represented in the new generated data. Cycling {is} due to large gradients. While large gradients are needed to escape local minima, thereby avoiding mode collapse, to prevent the cycling induced by large gradients, one needs stabilization. 
Existing approaches to train GANs address separately either cycling or mode collapse, but not both simultaneously.
Performing a {parallel independent} \cite{LupoPasini2021} training of Conditional GANs (CGANs) \cite{mirza_conditional_2014, yang2018diversitysensitive, Zhou2020, R1, miyato2018cgans, DBLP:journals/corr/abs-1912-04216} { that assigns different classes to different processes has been recently showed to} reduce the chances of mode collapse, but does not address cycling. Wasserstein GANs (WGANs) \cite{Heusel2020} address cycling, but do not address mode collapse. 

\textit{To simultaneously address cycling and mode collapse in the {parallel independent} training of GANs, we use the Wasserstein metric to define the cost functions associated with the discriminator and the generator, and adapt WGANs to a conditional variant \cite{LupoPasini2021}, where the data label is used as additional input to the generator in conjunction to the white noise. From now on, we refer to this variant as Wasserstein Conditional GANs (W-CGANs). } 
{ In situations where the data is characterized by a large number of data classes, our approach fully takes advantage of high performance computing (HPC) resources because the number of processes engaged in the training of GANs scales with the number of classes.}
Numerical results performed on CIFAR10 \cite{cifar10}, CIFAR100 \cite{cifar100} and ImageNet1k \cite{imagenet} show that W-CGANs stabilize the {parallel} training and lead to the production of better images with respect to past results obtained with the {parallel} training on deep convolutional GANs (DC-GANs) \cite{LupoPasini2021}. The performance of W-CGANs with respect to DC-CGANs is validated both in quantitative terms using performance metrics as well as by visual inspection. 

\section{Background on GANs}

In the context of GANs, original data and new generated data are described by two probability distributions. 
Given two probability distributions $p$ and $q$ defined on a metric space $\mathcal{X}$,
the Kullback–Leibler divergence (KL) \cite{KL} 
\begin{equation}
    KL(p,q) = \int_{\mathcal{X}} p(x) \log\bigg( \frac{p(x)}{q(x)} \bigg) dx,
\end{equation}
and its symmetrization, the Jensen–Shannon divergence (JS) \cite{Lin1991} 
\begin{equation}
    JS(p,q) = \frac{1}{2}KL(p,q) + \frac{1}{2} KL(q,p)
\end{equation}
measure the distance between the probability distributions $p$ and $q$ by computing the pointwise discrepancy in the values attained.
The Wasserstein distance between $p$ and $q$ is defined as 
\begin{equation}
    W(p,q) = \sup_{Lip(f)\le 1} \int_{\mathcal{X}} f(x)\big[ p(x) - q(x) \big] dx,
\label{wasserstein}
\end{equation}
where $Lip(f)\le 1$ represents the family of Lipschitz functions defined on $\mathcal{X}$ with Lipschitz constant less or equal to one. 
KL divergence and JS divergence attain indefinitely large values for any situations where the peaks of the two distributions do not overlap, and their value abruptly drops to zero only when the peaks of the distributions are located at the same point. In comparison, the Wasserstein metric represents a more informative estimate for measuring the distance between two probability distributions than KL divergence and JS divergence, since it measures the distance between the two expected values of the probabilities and suggests an update of the two probabilities to reduce that distance \cite{kl_wasserstein_comparison}. 

In order to contextualize the use of the Wasserstein metric for CGANs, we define the following input and output spaces, each with an associated probability distribution:
\begin{itemize}
    \item $Z$ is a noise space used to seed the generative model. $Z = \mathbb{R}^{d_Z}$ , where $d_Z$ is a hyperparameter. Values $\mathbf{z} \in Z$ are sampled from a noise distribution $p_\mathbf{z}(\mathbf{z})$. In our experiments $p_\mathbf{z}$ is a white noise distribution.
    \item $Y$ is an embedding space used to condition the generative model on additional external information, drawn from the training data. $Y = \mathbb{R}^{d_Y}$ , where $d_Y$ is a hyperparameter. Using conditional information provided in the training data, we define a density model $p_\mathbf{y}(\mathbf{y})$.
\item $X$ is the data space which represents an image output
from the generator or input to the discriminator. In our application the data are colored face images. Values
are normalized pixel values: $X = [0,1]^W \times
C$, where $W$ represents {the number of pixels in the}
images, and $C$ is the set of distinct color channels in
the input images. Using the images in the training data
and their associated conditional data, we can define
a density model $p_{\text{data}}(\mathbf{x})$ of face images. This is
exactly the density model we wish to replicate.
\end{itemize}
We now define two functions:
\begin{itemize}
\item $G: Z \times Y \rightarrow X$ is the conditional generative model (or generator), which accepts noise data $\mathbf{z}\in Z$ and produces an image $\mathbf{x}\in X$ conditional to the external information $\mathbf{y}\in Y$.
\item $D:X \rightarrow [0, 1]$ is the discriminative model (or
discriminator), which accepts an image $\mathbf{x}$ and condition
$\mathbf{y}$ and predicts the probability under condition $\mathbf{y}$ that $\mathbf{x}$
came from the empirical data distribution rather than
from the generative model.
\end{itemize}

The goal of the discriminator is to maximize the discrepancy between the data distribution and the probability distribution created by the generator $G$. For W-CGANs this means that the discriminator aims at maximizing the distance between the expected value of the probability distribution associated with the original data $\mathbb{E}_{\mathbf{x}\sim p_{\text{data}}} D(\mathbf{x})$ and the expected value of the probability distribution associated with fake data $\mathbb{E}_{\mathbf{z}} D(G(\mathbf{z}))$. 

In order to use the Wasserstein metric to define the cost function of the discriminator, we choose $D(\mathbf{x})$ to play the role of the function $f$ in Equation \eqref{wasserstein}, and we substitute the two probabilities $p$ and $q$ with $p_{\text{data}}: \mathcal{X}\rightarrow [0,1]$ and $p_{\text{model}}: Z \rightarrow [0,1]$.
Using the definition of expected value of $D$ under the probability distribution $p_{\text{data}}$ and $p_{\text{model}}$ as follows
\begin{equation}
\mathbb{E}_{\mathbf{x}\sim p_{\text{data}}} D(\mathbf{x}) = \int_{\mathcal{X}}D(\mathbf{x})
p_{\text{data}}(\mathbf{x})d\mathbf{x}
\end{equation}
\begin{equation} \mathbb{E}_{\mathbf{z}\sim p_{\text{model}}} D(G(\mathbf{z})) = \int_{\mathcal{Z}}D(G(\mathbf{z})) 
p_{\text{model}}(\mathbf{z})d\mathbf{z},
\end{equation}
we define the cost function for the discriminator as
\begin{equation}
    J^{(D)}_{Wasserstein}(\boldsymbol{\theta}^{(D)}, \boldsymbol{\theta}^{(G)}) = \sup_{\boldsymbol{\theta}^{(D)}} \bigg [\mathbb{E}_{\mathbf{x}\sim p_{\text{data}}} D(\mathbf{x}) - \mathbb{E}_{\mathbf{z}} D(G(\mathbf{z})) \bigg ].
    \label{cost_discriminator_wgan}
\end{equation}

The generator $G$ is parametrized with $\boldsymbol{\theta}^{(G)}$ and we define its cost function as
\begin{equation}
    J^{(G)}_{Wasserstein}(\boldsymbol{\theta}^{(D)}, \boldsymbol{\theta}^{(G)})  = \sup_{\boldsymbol{\theta}^{(G)}} \bigg [\mathbb{E}_{\mathbf{z}} D(G(\mathbf{z})) \bigg ].
    \label{cost_generator_wgan}
\end{equation}
Computing the Wasserstein metric in \eqref{cost_discriminator_wgan} and \eqref{cost_generator_wgan} is allowed by the fact neural networks used to model discriminator and generator are mathematically represented as Lipschitz functions \cite{Lipschitz}. 
Contextualizing the difference between the Wasserstein metric and the KL divergence and JS divergence in the case of CGANs, {we can} explain why the Wasserstein metric reduces the occurrence of cycle during the training of CGANs with respect to situations where the training of CGANs is driven by the minimization of the KL divergence. Indeed, the KL divergence and JS divergence can be reduced by keeping the expected value of the probability distribution fixed, and just augmenting the variance. However, by doing so, the characteristics of the points sampled by the generator (new data) may not necessarily change significantly. On the {other side}, the minimization of the Wasserstein metric forces the probability distribution associated with the new generated data to shift in order to have the expected value $\mathbb{E}_{\mathbf{z}} D(G(\mathbf{z}))$ get closer to $\mathbb{E}_{\mathbf{x}\sim p_{\text{data}}} D(\mathbf{x})$. Therefore, it is less likely that new data sampled at later iterations will resemble data generated at earlier iterations. 

\begin{figure}
\center
\includegraphics[width=\textwidth]{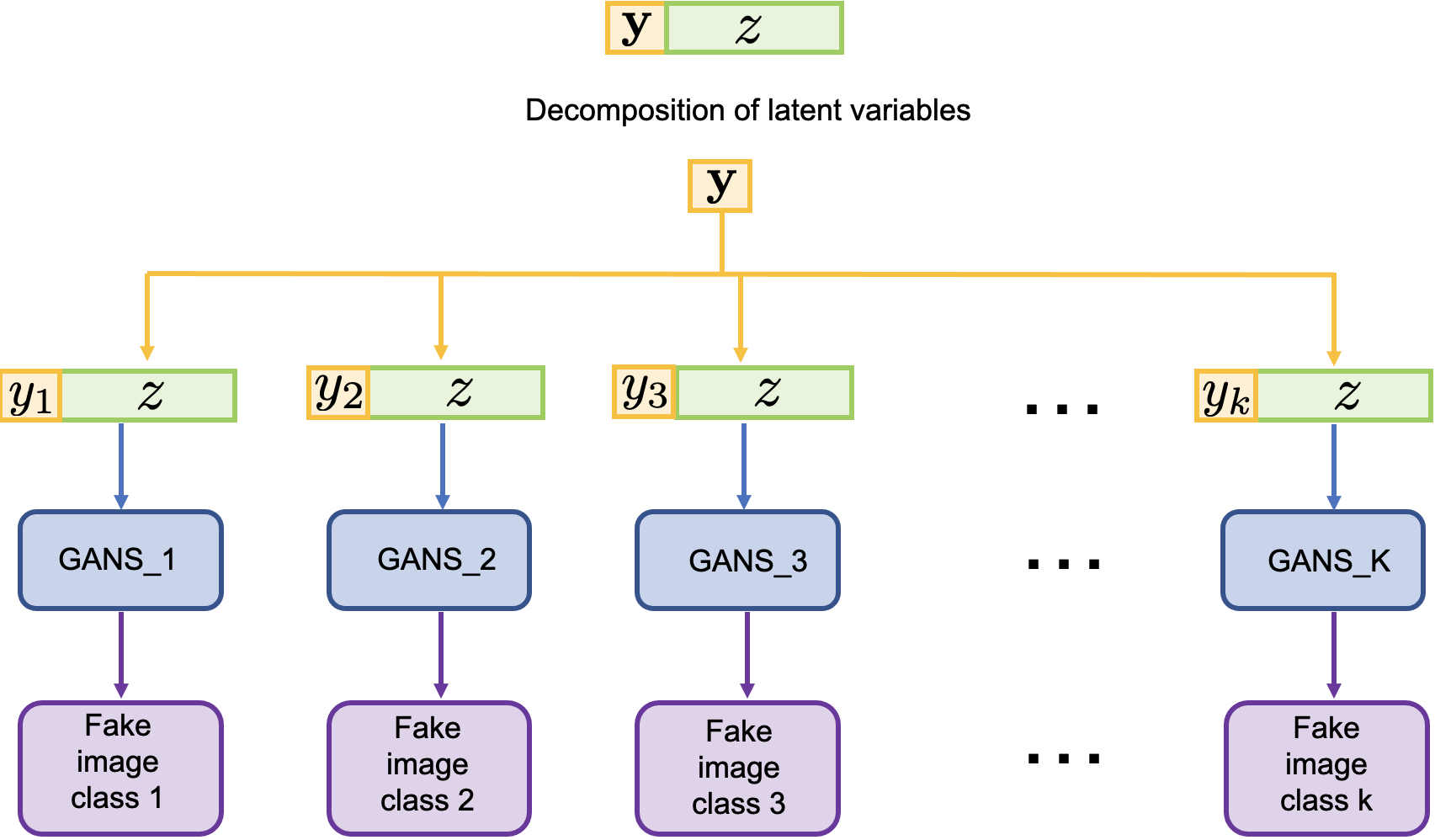}
\caption{Illustration of parallel W-CGANs.}
\label{parallel_gans_picture}
\end{figure}

Our {parallel} approach to train W-CGANs relies on the equality 
\begin{equation}
    p_\text{model}(\mathbf{x},\boldsymbol{\theta}) = \sum_{k=1}^K p_\text{model}(\mathbf{x},\boldsymbol{\theta}\lvert \mathbf{y}_k)p_\mathbf{y}(\mathbf{y}_k)
\end{equation}
that allows {to parallelize the computation} of each term $p_\text{model}(\mathbf{x},\mathbf{y}_k)$ by training $K$ {parallel} W-CGANs, each one per class, and then we combine the results at the end of each training to yield $p_\text{model}(\mathbf{x})$. The numerical examples presented in this paper are characterized by a one-to-one mapping between $\mathbf{y}_k$ and the labels in the image dataset. The advantage of our approach consists in the fact that all the $K$ {parallel} W-CGANs can be trained concurrently and independently of each other. If the complexity representation of the objects in each category is comparable, the training time for each {parallel} W-CGANs model is approximately the same, which in turn translates into promising performance in terms of weak scalability. An illustration that describe the distribution of W-CGANs is provided in Figure \ref{parallel_gans_picture}.

{ The parameters $\boldsymbol{\theta}$ for each replica of the W-CGAN model are updated independently using Adam. When the trained model is deployed, a random number generator provides the white noise and the label of the object whose image has to be generated. The randomly selected label determines which W-CGAN pair to call, and the white noise is passed to the selected W-CGAN pair to generate a new fake image for the specific object category associated with the label. Our parallel approach confines each model replica to be trained on data associated with a single label. In this sense, it differs from ensemble learning, where each model replica still has to span the entire dataset in a round-robin fashion. Ensemble learning can benefit the W-CGANs training in situations where interpolating across different classes is reasonable. However, image classification problems addressed in this work do not need interpolation (e.g., interpolating an image of a dog with an image of a car produces {clearly} unrealistic images). Therefore, parallel independent training is more appropriate than ensemble learning as it better accelerates the training by significantly reducing the amount of data processed by each GANs replica. Moreover, the partition of the data according to the classes facilitates our approach to scale, as confirmed by the weak scaling tests presented at the end of numerical section.}

\section{Numerical results}

In this section we present numerical results using CIFAR10, CIFAR100 and ImageNet1k benchmark image datasets to compare the performance of W-CGANs with DC-CGANs. The benchmark datasets CIFAR10 and CIFAR100 have the same total number of images, and images of both datasets have the same resolution, but the number of classes represented in the two datasets is different. Specifically, CIFAR10 has more images per class than CIFAR100. Since the GANs training is data intensive, the fact that CIFAR100 has fewer images per class results into a more challenging task for the model to produce good quality images.
The specifics of the neural networks used to model generator and discriminator is provided in Tables \ref{GANs_generator} and \ref{GANs_discriminator}. { Although the architectures of generator and discriminator can be tuned using hyperparameter optimization, this goes beyond the scope of this work since we aim at improving the performance of GANs for a fixed architecture. }

The training is performed using the optimizer Adam and a learning rate of $2\mathrm{e}-4$, and a total number of 1,000 epochs. 
The comparison between {parallel} DC-CGANs and {parallel} W-CGANs is performed on a quantitative level by measuring the Inception Score (IS) \cite{barratt_note_2018} and the Fréchet Inception Distance (FID) \cite{Heusel2020}. The IS takes a list of images and returns a single floating point number, the score, which is a measure of how realistic a GAN's output is. IS is an automatic alternative to having humans grade the quality of images.
The score measures two things simultaneously:
the image variety (e.g., each image is a different breed of dog), and whether each image distinctly looks like a real object. If both things are true, the score will be high. If either or both are false, the score will be low.
The lowest score possible is zero. Mathematically the highest possible score is infinity, although in practice a finite ceiling is imposed \cite{barratt_note_2018}.
Unlike IS, which evaluates only the distribution of generated images, the FID compares the distribution of generated images with the distribution of real images that were used to train the generator. Lower values of FID correspond to the distribution of generated images approaching the distribution of real images, and this is interpreted as an improvement of the generator in creating more realistic images.

\begin{table}
\centering
\begin{tabular}{|c|c|c|c|c|c|}
\hline
 \multicolumn{6}{|c|}{Generator} \\
\hline
Layer              & Input dim & Output dim & Kernel size & stride & padding\\
\hline
Input & 100 & 8192 & / & / & \\
\hline
\multicolumn{6}{|c|}{LeakyReLU(slope = 0.2, inplace=True)} \\
\hline
\multicolumn{6}{|c|}{Resizing} \\
\hline
\multicolumn{6}{|c|}{Batch normalization(epsilon = 1e-5 , momentum = 0.1)} \\
\hline
\multicolumn{6}{|c|}{Upsample(scale factor = 2)} \\
\hline
Conv.1            & 128 & 128 & 3 & 1 & 1\\
\hline
\multicolumn{6}{|c|}{Batch normalization(epsilon = 0.8 , momentum = 0.1)} \\
\hline
\multicolumn{6}{|c|}{LeakyReLU(slope = 0.2, inplace=True)} \\
\hline
\multicolumn{6}{|c|}{Upsample(scale factor = 2)} \\
\hline
Conv.2            & 128 & 64 & 3 & 1 & 1\\
\hline
\multicolumn{6}{|c|}{Batch normalization(epsilon = 0.8 , momentum = 0.1)} \\
\hline
\multicolumn{6}{|c|}{leakyReLU(slope = 0.2, inplace=True)} \\
\hline
Conv.3            & 64 & 1 or 3  & 3 & 1 & 1\\
\hline
\multicolumn{6}{|c|}{Tanh} \\
\hline
\end{tabular}
\caption{Architecture of the generator.}
\label{GANs_generator}
\end{table}

\begin{table}
\centering
\begin{tabular}{|c|c|c|c|c|c|}
\hline
 \multicolumn{6}{|c|}{Discriminator} \\
\hline
Layer              & In. dim & Out. dim & Kernel size & stride & padding\\
\hline
Conv.1            & 1 or 3 & 16 & 3 & 2 & 1\\
\hline
\multicolumn{6}{|c|}{leakyReLU(slope = 0.2, inplace=True)} \\
\hline
\multicolumn{6}{|c|}{Dropout(0.25)} \\
\hline
Conv.2            & 16 & 32 & 3 & 2 & 1\\
\hline
\multicolumn{6}{|c|}{LeakyReLU(slope = 0.2, inplace=True)} \\
\hline
\multicolumn{6}{|c|}{Dropout(0.25)} \\
\hline
\multicolumn{6}{|c|}{Batch normalization(epsilon = 0.8 , momentum = 0.1)} \\
\hline
Conv.3            & 32 & 64 &  3 & 2 & 1\\
\hline
\multicolumn{6}{|c|}{leakyReLU(slope = 0.2, inplace=True)} \\
\hline
\multicolumn{6}{|c|}{Dropout(0.25)} \\
\hline
\multicolumn{6}{|c|}{Batch normalization(epsilon = 0.8 , momentum = 0.1)} \\
\hline
Conv.4            & 64 & 128 &  3 & 2 & 1\\
\hline
\multicolumn{6}{|c|}{leakyReLU(slope = 0.2, inplace=True)} \\
\hline
\multicolumn{6}{|c|}{Dropout(0.25)} \\
\hline
\multicolumn{6}{|c|}{Batch normalization(epsilon = 0.8 , momentum = 0.1)} \\
\hline
output & 2048 & 1 &  / & / & /\\
\hline
\multicolumn{6}{|c|}{Sigmoid(for DC-CGANs), no activation function for W-CGANs} \\
\hline
\end{tabular}
\caption{Architecture of the discriminator.}
\label{GANs_discriminator}
\end{table}

\subsection{Hardware description}
The numerical experiments are performed using Summit \cite{summit}, a supercomputer at the Oak Ridge Leadership Computing Facility (OLCF) at Oak Ridge National Laboratory. Summit has a hybrid architecture, and each node contains two IBM POWER9 CPUs and six NVIDIA Volta GPUs all connected together with NVIDIA’s high-speed NVLink. Each node has over half a terabyte of coherent memory (high bandwidth memory + DDR4) addressable by all CPUs and GPUs plus 1.6 TB of non-volatile memory (NVMe) storage that can be used as a burst buffer or as extended memory. To provide a high rate of communication and I/O throughput, the nodes are connected in a non-blocking fat-tree using a dual-rail Mellanox EDR InfiniBand interconnect.

\subsection{Software description}
The numerical experiments are performed using \texttt{Python3.7} with \texttt{PyTorch v1.3.1} package \cite{paszke_pytorch_2019} 
for autodifferentiation to train the DL models with the use of GPUs, and the \texttt{torch.nn.parallel.DistributedDataParallel} tool is used for parallel computing to coordinate the different W-CGAN model replicas.

\subsection{CIFAR10}
The training portion of the CIFAR10 dataset \cite{cifar10} consists of 50,000 32x32 color images in 10 classes, with 5,000 images per class. The classes represented in the dataset are airplanes, automobiles, birds, cats, deers, fogs, frogs, horses, ships, and trucks. 

\begin{figure}
\resizebox{\columnwidth}{!}{
\begin{tabular}{|c|c|c|c|c|c|}
      \hline
      \includegraphics[width=10mm]{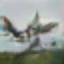} &
      \includegraphics[width=10mm]{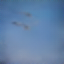} &
      \includegraphics[width=10mm]{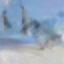} &     
      \includegraphics[width=10mm]{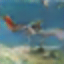} & 
      \includegraphics[width=10mm]{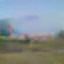} &    
      \includegraphics[width=10mm]{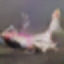} \\        
      \includegraphics[width=10mm]{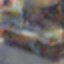} &
      \includegraphics[width=10mm]{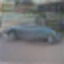} &          
      \includegraphics[width=10mm]{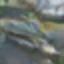} &
      \includegraphics[width=10mm]{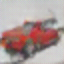} &  
      \includegraphics[width=10mm]{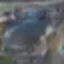} &     
      \includegraphics[width=10mm]{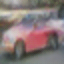} \\   
      \includegraphics[width=10mm]{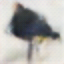} &   
      \includegraphics[width=10mm]{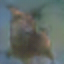} &   
      \includegraphics[width=10mm]{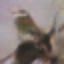} &      
      \includegraphics[width=10mm]{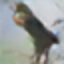} &      
      \includegraphics[width=10mm]{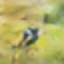} &   
      \includegraphics[width=10mm]{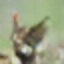} \\   
      \includegraphics[width=10mm]{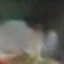} &
      \includegraphics[width=10mm]{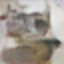} &    
      \includegraphics[width=10mm]{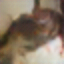} &         
      \includegraphics[width=10mm]{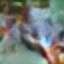} &      
      \includegraphics[width=10mm]{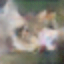} &    
      \includegraphics[width=10mm]{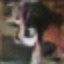}         
      \\      
      \includegraphics[width=10mm]{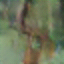} &
      \includegraphics[width=10mm]{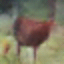} &
      \includegraphics[width=10mm]{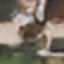} &
      \includegraphics[width=10mm]{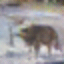} &
      \includegraphics[width=10mm]{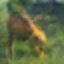} &    
      \includegraphics[width=10mm]{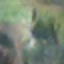} \\         
      \includegraphics[width=10mm]{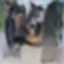} &
      \includegraphics[width=10mm]{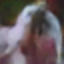} &
      \includegraphics[width=10mm]{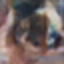} &
      \includegraphics[width=10mm]{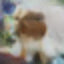} &      
      \includegraphics[width=10mm]{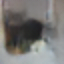} &
      \includegraphics[width=10mm]{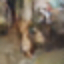}        
      \\
      \includegraphics[width=10mm]{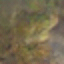} &      
      \includegraphics[width=10mm]{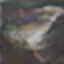} &
      \includegraphics[width=10mm]{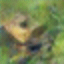} &
      \includegraphics[width=10mm]{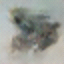} &            
      \includegraphics[width=10mm]{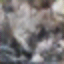} &  
      \includegraphics[width=10mm]{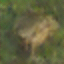} \\
      \includegraphics[width=10mm]{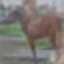} &
      \includegraphics[width=10mm]{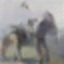} &
      \includegraphics[width=10mm]{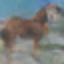} &
      \includegraphics[width=10mm]{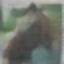} &   
      \includegraphics[width=10mm]{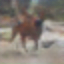} &         
      \includegraphics[width=10mm]{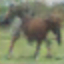} \\           
      \includegraphics[width=10mm]{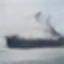} &
      \includegraphics[width=10mm]{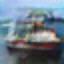} &   
      \includegraphics[width=10mm]{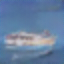} &     
      \includegraphics[width=10mm]{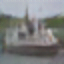} &           
      \includegraphics[width=10mm]{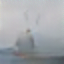} &
      \includegraphics[width=10mm]{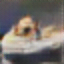} \\           
      \includegraphics[width=10mm]{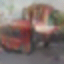} &
      \includegraphics[width=10mm]{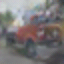} &
      \includegraphics[width=10mm]{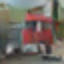} &
      \includegraphics[width=10mm]{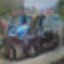} &     
      \includegraphics[width=10mm]{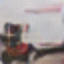} &     
      \includegraphics[width=10mm]{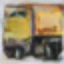} \\                
      \hline
\end{tabular}
}
\caption{Fake images generated by {parallel} DC-CGANs trained on the CIFAR10 dataset.}
\label{CIFAR10_Adam_Distributed}
\end{figure}

\begin{figure}
   \centering
\resizebox{\columnwidth}{!}{
\begin{tabular}{|c|c|c|c|c|c|}
      \hline
      \includegraphics[width=10mm]{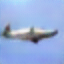} &        
      \includegraphics[width=10mm]{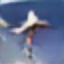} &        
      \includegraphics[width=10mm]{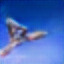} &        
      \includegraphics[width=10mm]{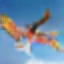} &        
      \includegraphics[width=10mm]{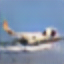} &        
      \includegraphics[width=10mm]{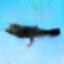}       
      \\
      \includegraphics[width=10mm]{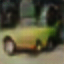} &        
      \includegraphics[width=10mm]{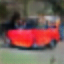} &        
      \includegraphics[width=10mm]{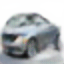} &         
      \includegraphics[width=10mm]{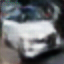} &      
      \includegraphics[width=10mm]{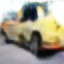}&          
      \includegraphics[width=10mm]{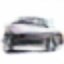}            
      \\
      \includegraphics[width=10mm]{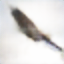} &     
      \includegraphics[width=10mm]{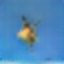} &    
      \includegraphics[width=10mm]{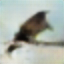} &   
      \includegraphics[width=10mm]{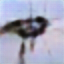} &   
      \includegraphics[width=10mm]{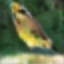} &  
      \includegraphics[width=10mm]{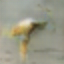}              
      \\
      \includegraphics[width=10mm]{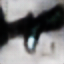} &         
      \includegraphics[width=10mm]{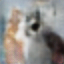} &         
      \includegraphics[width=10mm]{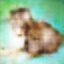} &     
      \includegraphics[width=10mm]{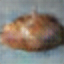} &   
      \includegraphics[width=10mm]{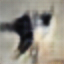} &      
      \includegraphics[width=10mm]{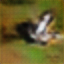}  
      \\
      \includegraphics[width=10mm]{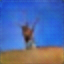} &     
      \includegraphics[width=10mm]{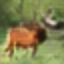} &        
      \includegraphics[width=10mm]{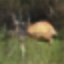} &        
      \includegraphics[width=10mm]{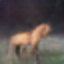} &        
      \includegraphics[width=10mm]{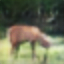} &        
      \includegraphics[width=10mm]{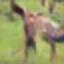}        
      \\
      \includegraphics[width=10mm]{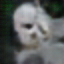} &  
      \includegraphics[width=10mm]{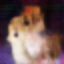} &       
      \includegraphics[width=10mm]{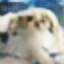} &  
      \includegraphics[width=10mm]{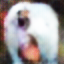} &   
      \includegraphics[width=10mm]{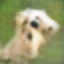} &        
      \includegraphics[width=10mm]{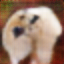}      
      \\
      \includegraphics[width=10mm]{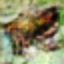} &    
      \includegraphics[width=10mm]{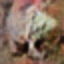} &  
      \includegraphics[width=10mm]{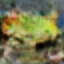} &  
      \includegraphics[width=10mm]{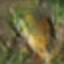} &  
      \includegraphics[width=10mm]{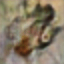} &   
      \includegraphics[width=10mm]{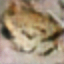}       
      \\
      \includegraphics[width=10mm]{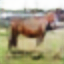} &   
      \includegraphics[width=10mm]{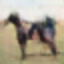} &    
      \includegraphics[width=10mm]{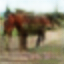} &  
      \includegraphics[width=10mm]{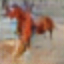} &  
      \includegraphics[width=10mm]{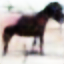} &       
      \includegraphics[width=10mm]{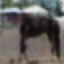}    
      \\
      \includegraphics[width=10mm]{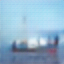} &         
      \includegraphics[width=10mm]{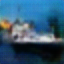} &            
      \includegraphics[width=10mm]{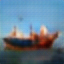} &           
      \includegraphics[width=10mm]{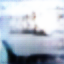} &      
      \includegraphics[width=10mm]{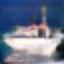} &
      \includegraphics[width=10mm]{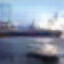} \\      
      \includegraphics[width=10mm]{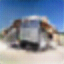} &
      \includegraphics[width=10mm]{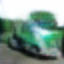} &
      \includegraphics[width=10mm]{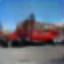} &     
      \includegraphics[width=10mm]{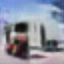} & 
      \includegraphics[width=10mm]{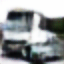} &    
      \includegraphics[width=10mm]{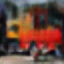} \\ 
      \hline
\end{tabular}
}
\caption{Fake images generated by {parallel} W-CGANs trained on the CIFAR10 dataset.}
\label{CIFAR10_WGANs}
\end{figure}

A comparison in quantitative terms between {parallel} DC-CGANs and the {parallel} W-CGANs is shown in Table \ref{cifar10_table} where the performance of the models is measured in terms of IS and FID, and W-CGANs outperform DC-CGANs with respect to both indices. A visual comparison between the images produced by DC-CGANs in Figure \ref{CIFAR10_Adam_Distributed} and the images produced by W-CGANs in Figure \ref{CIFAR10_WGANs} shows that W-CGANs succeeds produces objects with more refined contours, and the objects are much easier to recognize with respect to the object class they attempt at sampling.
Weak scaling plot for DC-CGANs and W-CGANs trained on the CIFAR10 dataset is presented on the left side of Figure \ref{scalability_pic} by reporting the runtime of the slowest process. The independence of the generator-discriminator pairs allows the code to scale with the number of processors that take care of separate data classes. 
The average GPU utilization is 87.5\% with a standard deviation of 2.2\%, and the memory utilization is 7,839 mebibytes (MiB).

\begin{table}
\centering
\begin{tabular}{|c|c|c|}
\cline{1-3}
 & IS & FID \\
 \cline{1-3}
{Parallel} DC-CGANs & 6.43 & 9.41\\ \cline{1-3}
{Parallel} W-CGANs & 7.43 & 8.53\\ \cline{1-3}
\end{tabular}
\caption{Inception score (IS) and Fréchet Inception Distance (FID) for the training of {parallel} DC-CGANs and {parallel} W-CGANs on CIFAR10. }
\label{cifar10_table}
\end{table}

\subsection{CIFAR100}
The training portion of the CIFAR100 dataset \cite{cifar100} consists of 50,000 32x32 color images in 100 classes, with 500 images per class. We refer the reader to \cite{cifar100} for details about the objects represented in each class. 

\begin{figure}
\centering
\begin{tabular}{|c|c|c|c|c|c|c|c|}
      \hline
      \includegraphics[width=10mm]{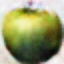} &
      \includegraphics[width=10mm]{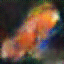} &   
      \includegraphics[width=10mm]{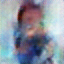} &      
      \includegraphics[width=10mm]{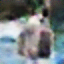} &    
      \includegraphics[width=10mm]{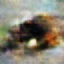} &    
      \includegraphics[width=10mm]{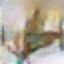} &    
      \includegraphics[width=10mm]{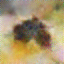} &      
      \includegraphics[width=10mm]{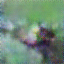} \\      
      \includegraphics[width=10mm]{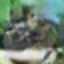} &      
      \includegraphics[width=10mm]{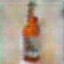} &
      \includegraphics[width=10mm]{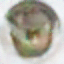} &
      \includegraphics[width=10mm]{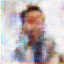} &  
      \includegraphics[width=10mm]{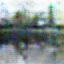} &    
      \includegraphics[width=10mm]{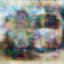} &    
      \includegraphics[width=10mm]{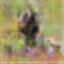} &   
      \includegraphics[width=10mm]{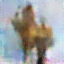} \\      
      \includegraphics[width=10mm]{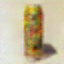} &
      \includegraphics[width=10mm]{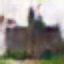} &
      \includegraphics[width=10mm]{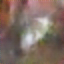} &      
      \includegraphics[width=10mm]{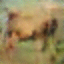} &     
      \includegraphics[width=10mm]{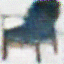} &
      \includegraphics[width=10mm]{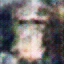} &      
      \includegraphics[width=10mm]{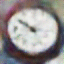} &
      \includegraphics[width=10mm]{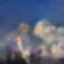} \\
      \includegraphics[width=10mm]{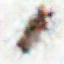} &      
      \includegraphics[width=10mm]{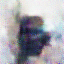} &     
      \includegraphics[width=10mm]{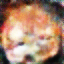} &   
      \includegraphics[width=10mm]{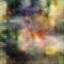} &      
      \includegraphics[width=10mm]{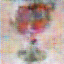} &
      \includegraphics[width=10mm]{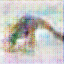} &          
      \includegraphics[width=10mm]{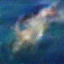} & 
      \includegraphics[width=10mm]{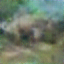} \\    
      \includegraphics[width=10mm]{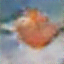} &     
      \includegraphics[width=10mm]{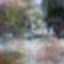} &      
      \includegraphics[width=10mm]{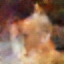} &      
      \includegraphics[width=10mm]{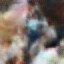} &  
      \includegraphics[width=10mm]{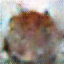} &   
      \includegraphics[width=10mm]{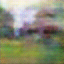} &      
      \includegraphics[width=10mm]{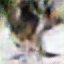} &      
      \includegraphics[width=10mm]{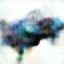} \\    
      \includegraphics[width=10mm]{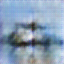} &      
      \includegraphics[width=10mm]{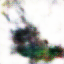} &
      \includegraphics[width=10mm]{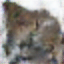} &      
      \includegraphics[width=10mm]{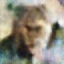} &     
      \includegraphics[width=10mm]{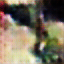} &      
      \includegraphics[width=10mm]{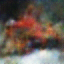} &     
      \includegraphics[width=10mm]{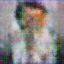} &     
      \includegraphics[width=10mm]{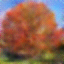} \\
      \includegraphics[width=10mm]{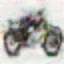} &
      \includegraphics[width=10mm]{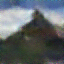} &
      \includegraphics[width=10mm]{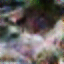} &      
      \includegraphics[width=10mm]{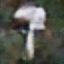} &
      \includegraphics[width=10mm]{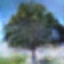} &
      \includegraphics[width=10mm]{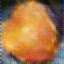} &      
      \includegraphics[width=10mm]{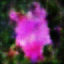} &
      \includegraphics[width=10mm]{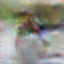} \\      
      \includegraphics[width=10mm]{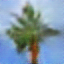} &
      \includegraphics[width=10mm]{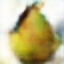} &
      \includegraphics[width=10mm]{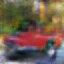} &
      \includegraphics[width=10mm]{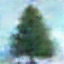} &
      \includegraphics[width=10mm]{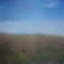} &
      \includegraphics[width=10mm]{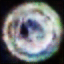} &
      \includegraphics[width=10mm]{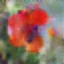} &      
      \includegraphics[width=10mm]{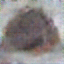} \\      
      \includegraphics[width=10mm]{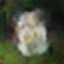} &      
      \includegraphics[width=10mm]{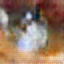} &      
      \includegraphics[width=10mm]{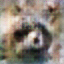} &      
      \includegraphics[width=10mm]{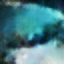} &      
      \includegraphics[width=10mm]{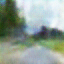} &
      \includegraphics[width=10mm]{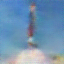} &      
      \includegraphics[width=10mm]{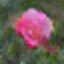} &      
      \includegraphics[width=10mm]{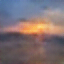} \\
      \includegraphics[width=10mm]{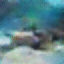} &     
      \includegraphics[width=10mm]{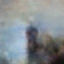} &     
      \includegraphics[width=10mm]{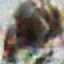} &      
      \includegraphics[width=10mm]{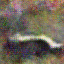} &  
      \includegraphics[width=10mm]{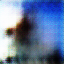} &      
      \includegraphics[width=10mm]{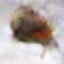} &
      \includegraphics[width=10mm]{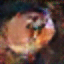} &      
      \includegraphics[width=10mm]{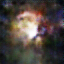} \\    
      \includegraphics[width=10mm]{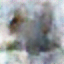} &      
      \includegraphics[width=10mm]{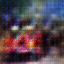} &      
      \includegraphics[width=10mm]{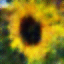} &
      \includegraphics[width=10mm]{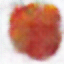} &      
      \includegraphics[width=10mm]{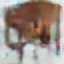} &
      \includegraphics[width=10mm]{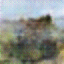} &    
      \includegraphics[width=10mm]{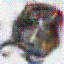} &      
      \includegraphics[width=10mm]{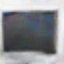} \\
      \includegraphics[width=10mm]{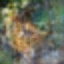} &      
      \includegraphics[width=10mm]{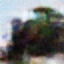} &  
      \includegraphics[width=10mm]{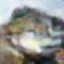} &      
      \includegraphics[width=10mm]{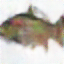} &
      \includegraphics[width=10mm]{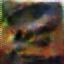} &      
      \includegraphics[width=10mm]{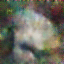} &      
     \includegraphics[width=10mm]{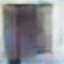} &  
      \includegraphics[width=10mm]{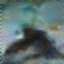} \\     
      \includegraphics[width=10mm]{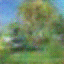} &  
      \includegraphics[width=10mm]{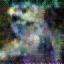} & 
      \includegraphics[width=10mm]{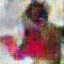} &   
      \includegraphics[width=10mm]{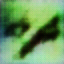}      
\end{tabular}
\caption{Fake images generated by {parallel} DC-CGANs trained on the CIFAR100 dataset. 
}
\label{CIFAR100_Adam_Distributed}
\end{figure}

\begin{figure}
   \centering
\begin{tabular}{|c|c|c|c|c|c|c|c|}
      \hline
      \includegraphics[width=10mm]{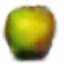} &
      \includegraphics[width=10mm]{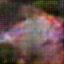} &      
      \includegraphics[width=10mm]{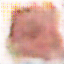} &            
      \includegraphics[width=10mm]{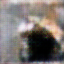} &     
      \includegraphics[width=10mm]{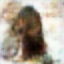} &           
     \includegraphics[width=10mm]{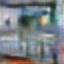} &           
     \includegraphics[width=10mm]{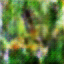} &    
      \includegraphics[width=10mm]{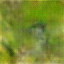} \\     
      \includegraphics[width=10mm]{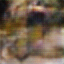} &      
     \includegraphics[width=10mm]{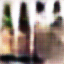} &
      \includegraphics[width=10mm]{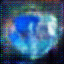} &      
      \includegraphics[width=10mm]{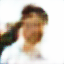} &   
      \includegraphics[width=10mm]{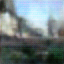} &      
      \includegraphics[width=10mm]{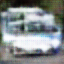} & 
      \includegraphics[width=10mm]{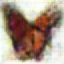} &   
      \includegraphics[width=10mm]{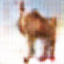} \\        
      \includegraphics[width=10mm]{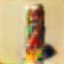} &       
      \includegraphics[width=10mm]{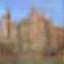} &
      \includegraphics[width=10mm]{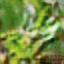} &   
      \includegraphics[width=10mm]{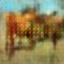} &         
      \includegraphics[width=10mm]{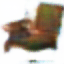} &
      \includegraphics[width=10mm]{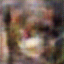} &      
      \includegraphics[width=10mm]{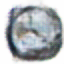} &
      \includegraphics[width=10mm]{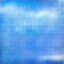} \\         
      \includegraphics[width=10mm]{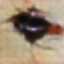} &
      \includegraphics[width=10mm]{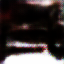} &
      \includegraphics[width=10mm]{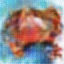} &    
      \includegraphics[width=10mm]{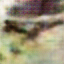} &   
      \includegraphics[width=10mm]{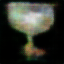} &         
      \includegraphics[width=10mm]{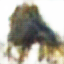} &      
      \includegraphics[width=10mm]{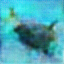} &        
      \includegraphics[width=10mm]{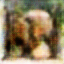} \\  
      \includegraphics[width=10mm]{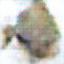} &         
      \includegraphics[width=10mm]{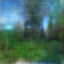} &              
      \includegraphics[width=10mm]{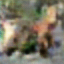} &   
      \includegraphics[width=10mm]{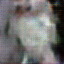} &              
      \includegraphics[width=10mm]{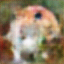} &       
      \includegraphics[width=10mm]{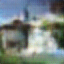} &              
      \includegraphics[width=10mm]{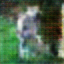} &           
      \includegraphics[width=10mm]{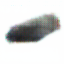} \\     
      \includegraphics[width=10mm]{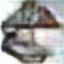} &              
      \includegraphics[width=10mm]{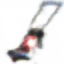} &
      \includegraphics[width=10mm]{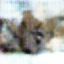} &      
      \includegraphics[width=10mm]{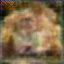} &           
      \includegraphics[width=10mm]{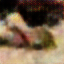} &            
      \includegraphics[width=10mm]{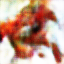} &      
      \includegraphics[width=10mm]{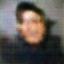} &            
      \includegraphics[width=10mm]{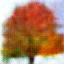} \\    
      \includegraphics[width=10mm]{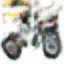} &  
      \includegraphics[width=10mm]{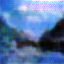} &  
      \includegraphics[width=10mm]{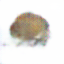} &        
      \includegraphics[width=10mm]{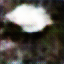} &              
      \includegraphics[width=10mm]{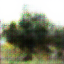} &
      \includegraphics[width=10mm]{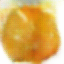} &      
      \includegraphics[width=10mm]{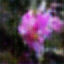} &
      \includegraphics[width=10mm]{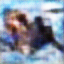} \\      
      \includegraphics[width=10mm]{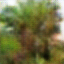} &
      \includegraphics[width=10mm]{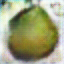} &      
      \includegraphics[width=10mm]{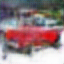} &
      \includegraphics[width=10mm]{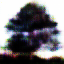} &      
      \includegraphics[width=10mm]{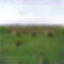} &
      \includegraphics[width=10mm]{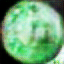} &
      \includegraphics[width=10mm]{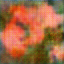} &
      \includegraphics[width=10mm]{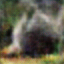} \\
      \includegraphics[width=10mm]{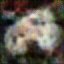} &      
      \includegraphics[width=10mm]{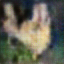} &      
      \includegraphics[width=10mm]{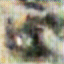} &    
      \includegraphics[width=10mm]{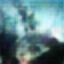} &  
      \includegraphics[width=10mm]{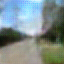} &        
      \includegraphics[width=10mm]{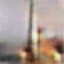} &
      \includegraphics[width=10mm]{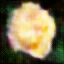} &           
      \includegraphics[width=10mm]{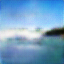} \\      
      \includegraphics[width=10mm]{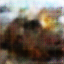} &
      \includegraphics[width=10mm]{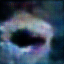} &      
      \includegraphics[width=10mm]{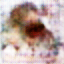} &            
     \includegraphics[width=10mm]{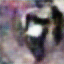} &
     \includegraphics[width=10mm]{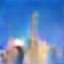} &
     \includegraphics[width=10mm]{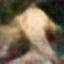} &    
     \includegraphics[width=10mm]{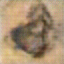} &    
     \includegraphics[width=10mm]{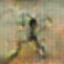} \\       
     \includegraphics[width=10mm]{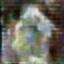} &       
     \includegraphics[width=10mm]{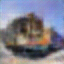} &       
      \includegraphics[width=10mm]{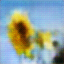} &
      \includegraphics[width=10mm]{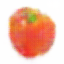} &
      \includegraphics[width=10mm]{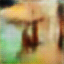} &  
      \includegraphics[width=10mm]{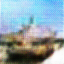} &     
      \includegraphics[width=10mm]{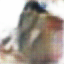} & 
      \includegraphics[width=10mm]{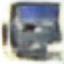} \\    
      \includegraphics[width=10mm]{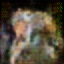} &      
      \includegraphics[width=10mm]{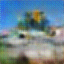} &          
      \includegraphics[width=10mm]{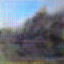} &      
      \includegraphics[width=10mm]{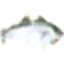} &     
      \includegraphics[width=10mm]{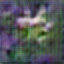} &         
      \includegraphics[width=10mm]{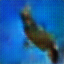} &           
      \includegraphics[width=10mm]{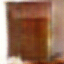} &
      \includegraphics[width=10mm]{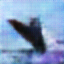} \\   
      \includegraphics[width=10mm]{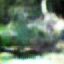} &           
      \includegraphics[width=10mm]{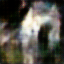} &   
     \includegraphics[width=10mm]{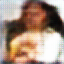} &         
     \includegraphics[width=10mm]{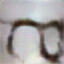}       
\end{tabular}
\caption{Fake images generated by {parallel} W-CGANs trained on CIFAR100. 
}
\label{CIFAR100_WGANs}
\end{figure}

The fact that CIFAR100 has fewer image data per class than CIFAR10 makes the GANs training more difficult, but the use of the Wasserstein metric stabilizes the {parallel} training of W-CGANs, which results into a higher IS score and lower FID score as shown in Table \ref{cifar100_table}. 
A visual comparison between the images produced by DC-CGANs in Figure \ref{CIFAR100_Adam_Distributed} and the images produced by W-CGANs in Figure \ref{CIFAR100_WGANs} shows that W-CGANs outperforms DC-CGANs on objects that are particularly complex to represent, such as houses surrounded by a garden (second row, third picture from the left), butterflies (second row, fifth picture from the left), and camels (second row, sixth picture from the left).

Weak scaling plot for DC-CGANs and W-CGANs trained on the CIFAR100 dataset is presented on the right side of Figure \ref{scalability_pic}, by reporting the runtime of the slowest process. Also in this case, the independence of the generator-discriminator pairs allows the code to scale with the number of processors that take care of separate data classes. 
The average GPU utilization is 93.8\% with a standard deviation of 0.7\%, and the memory utilization is 7,839 MiB.

\begin{table}
\centering
\begin{tabular}{|c|c|c|}
\cline{1-3}
 & IS & FID \\ \cline{1-3}
{Parallel} DC-CGANs & 6.61 & 9.23 \\ \cline{1-3}
{Parallel} W-CGANs & 6.93 & 8.92 \\ \cline{1-3}
\end{tabular}
\caption{Inception score (IS) and Fréchet Inception Distance (FID) for the training of {parallel} DC-CGANs and {parallel} W-CGANs on the CIFAR100 dataset.}
\label{cifar100_table}
\end{table}

\subsection{ImageNet1k}
The training portion of the ImageNet1k dataset \cite{imagenet} consists of a training set of 1.2 million colored images and a test set of 50,000 colored images in 1,000 classes, with 50 images per class. We refer the reader to \cite{cifar100} for details about the objects represented in each class. The resolution of the images varies across the samples, with an average pixel size equal to 482x418.

The comparison between DC-CGANs and W-CGANs in terms of IS score and FID score is shown in Table \ref{imagenet_table}, where W-CGANs shows an improvement with respect to DC-CGANs in terms of both figures of merits. 

\begin{figure}
\centering
\begin{tabular}{|c|c|c|c|c|c|}
      \hline
      \includegraphics[width=10mm]{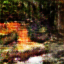} &
      \includegraphics[width=10mm]{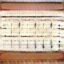} &    
      \includegraphics[width=10mm]{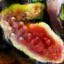} &    
     \includegraphics[width=10mm]{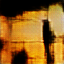} &
      \includegraphics[width=10mm]{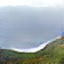} &
      \includegraphics[width=10mm]{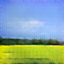} \\  
            \hline
      \includegraphics[width=10mm]{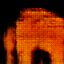}&
      \includegraphics[width=10mm]{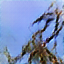} &
      \includegraphics[width=10mm]{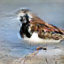} &
      \includegraphics[width=10mm]{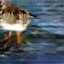} &
      \includegraphics[width=10mm]{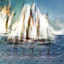} &
      \includegraphics[width=10mm]{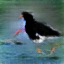} \\
            \hline
      \includegraphics[width=10mm]{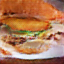} & 
      \includegraphics[width=10mm]{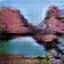} &
      \includegraphics[width=10mm]{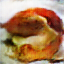} &     
     \includegraphics[width=10mm]{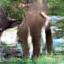} &     
      \includegraphics[width=10mm]{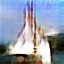} &
       \includegraphics[width=10mm]{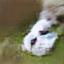} \\
            \hline
      \includegraphics[width=10mm]{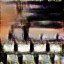} &
      \includegraphics[width=10mm]{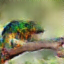} &
      \includegraphics[width=10mm]{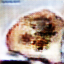} &
      \includegraphics[width=10mm]{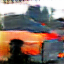} &
     \includegraphics[width=10mm]{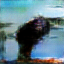} &
      \includegraphics[width=10mm]{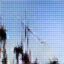}\\
            \hline
      \includegraphics[width=10mm]{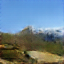} &
      \includegraphics[width=10mm]{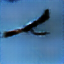} &
      \includegraphics[width=10mm]{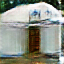} &
     \includegraphics[width=10mm]{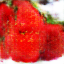} &      
      \includegraphics[width=10mm]{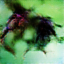} &
     \includegraphics[width=10mm]{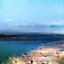} \\ 
     \hline
     \includegraphics[width=10mm]{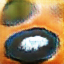} &
      \includegraphics[width=10mm]{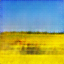} &
      \includegraphics[width=10mm]{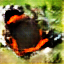} &
      \includegraphics[width=10mm]{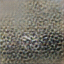}  &
      \includegraphics[width=10mm]{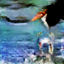} &
     \includegraphics[width=10mm]{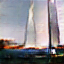} \\   
     \hline
     \includegraphics[width=10mm]{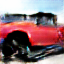} &
      \includegraphics[width=10mm]{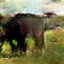} &
      \includegraphics[width=10mm]{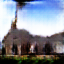} &
      \includegraphics[width=10mm]{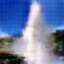} &
      \includegraphics[width=10mm]{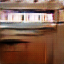} &
     \includegraphics[width=10mm]{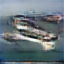}   \\         
\end{tabular}
\caption{Fake images generated by parallel DC-CGANs trained on ImageNet1k. 
}
\label{Imagenet_Adam_Distributed}
\end{figure}

\begin{figure}
\centering
\begin{tabular}{|c|c|c|c|c|c|}
      \hline
      \includegraphics[width=10mm]{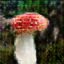} &
      \includegraphics[width=10mm]{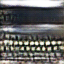} &    
      \includegraphics[width=10mm]{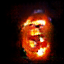} &    
     \includegraphics[width=10mm]{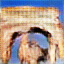} &
     \includegraphics[width=10mm]{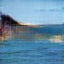} &
      \includegraphics[width=10mm]{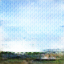} \\  
            \hline
      \includegraphics[width=10mm]{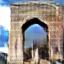}  &
      \includegraphics[width=10mm]{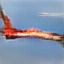} &
      \includegraphics[width=10mm]{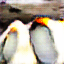} &
      \includegraphics[width=10mm]{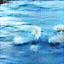} &
      \includegraphics[width=10mm]{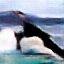} &
      \includegraphics[width=10mm]{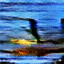} \\
            \hline
     \includegraphics[width=10mm]{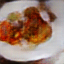} & 
     \includegraphics[width=10mm]{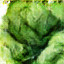} &
     \includegraphics[width=10mm]{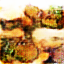} &
     \includegraphics[width=10mm]{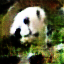} &     
     \includegraphics[width=10mm]{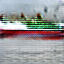} &     
       \includegraphics[width=10mm]{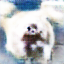} \\
            \hline
       \includegraphics[width=10mm]{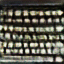}  &
      \includegraphics[width=10mm]{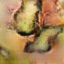} &
      \includegraphics[width=10mm]{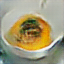} &
      \includegraphics[width=10mm]{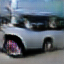} &
     \includegraphics[width=10mm]{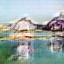} &
      \includegraphics[width=10mm]{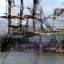}\\
            \hline
      \includegraphics[width=10mm]{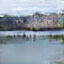} &
      \includegraphics[width=10mm]{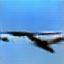}&
      \includegraphics[width=10mm]{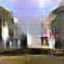}&
      \includegraphics[width=10mm]{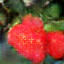} &      
      \includegraphics[width=10mm]{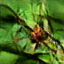}&
     \includegraphics[width=10mm]{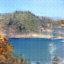} \\ 
     \hline
     \includegraphics[width=10mm]{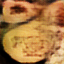}  &
      \includegraphics[width=10mm]{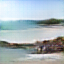} &
      \includegraphics[width=10mm]{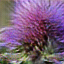} &
      \includegraphics[width=10mm]{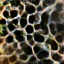} &
      \includegraphics[width=10mm]{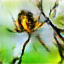} &   
    \includegraphics[width=10mm]{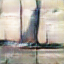}  \\   
     \hline
     \includegraphics[width=10mm]{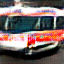}  &
     \includegraphics[width=10mm]{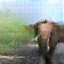}  &
         \includegraphics[width=10mm]{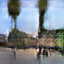} &
          \includegraphics[width=10mm]{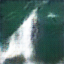} &
          \includegraphics[width=10mm]{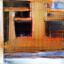} &
     \includegraphics[width=10mm]{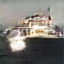}\\    
\end{tabular}
\caption{Fake images generated by parallel W-CGANs trained on ImageNet1k. 
}
\label{Imagenet_Adam_WGANs}
\end{figure}

A visual comparison between the images produced by DC-CGANs in Figure \ref{Imagenet_Adam_Distributed} and the images produced by W-CGANs in Figure \ref{Imagenet_Adam_WGANs} shows that W-CGANs outperforms DC-CGANs on objects that are particularly complex to represent, such as monuments (second row, first picture from the left), butterflies (second row, fifth picture from the left), and camels (second row, sixth picture from the left).

\begin{table}
\centering
\begin{tabular}{|c|c|c|}
\cline{1-3}
 & IS & FID \\ \cline{1-3}
{Parallel} DC-CGANs & 11.61 & 12.23 \\ \cline{1-3}
{Parallel} W-CGANs & 12.93 & 11.92 \\ \cline{1-3}
\end{tabular}
\caption{Inception score (IS) and Fréchet Inception Distance (FID) for the training of {parallel} DC-CGANs and {parallel} W-CGANs on the ImaheNet1k dataset.}
\label{imagenet_table}
\end{table}

\subsection{Scaling performance of parallel training of W-CGANs}
\begin{figure}
   \centering
\begin{tabular}{cc}
\includegraphics[width=0.5\textwidth]{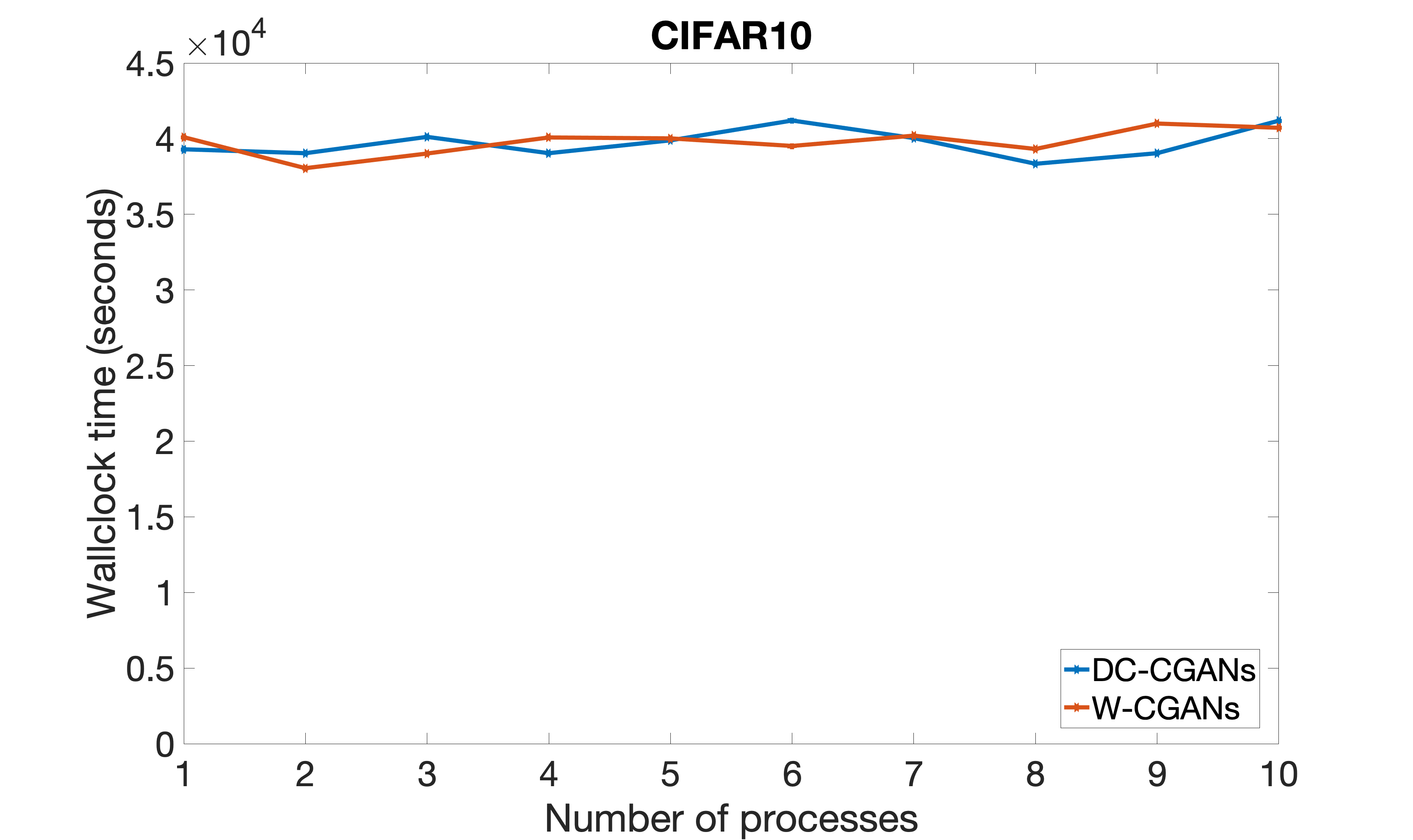}
\includegraphics[width=0.5\textwidth]{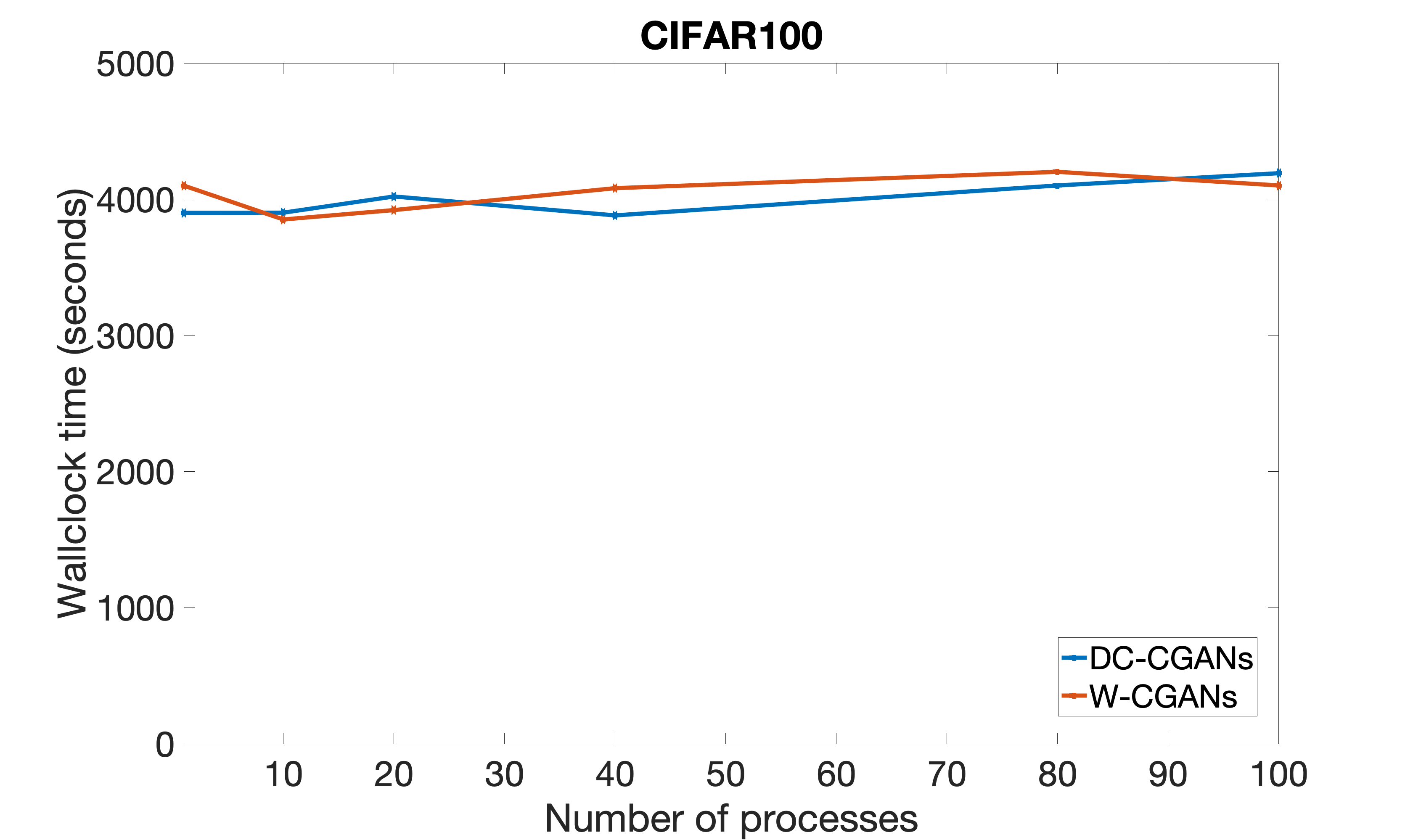}
\end{tabular}
    \caption{Results of weak scaling tests for CIFAR10, and CIFAR100. Wall-clock time measured in seconds is reported for the slowest processor.}
    \label{scalability_pic} 
\end{figure}

We tested the scalability of our parallel approach to train W-CGANs by running experiments on OLCF supercomputer Summit. We measured the wall-clock time needed by the parallel W-CGANs to complete the training as a function of the number of data classes (and thus processes). The parameters for the numerical optimization are the same as discussed before. The training of the model has been performed by distributing the computation through a one-to-one mapping between the  process and the data classes. Each process was mapped to two NVIDIA V100 GPUs, so that the neural networks for discriminator and generator for each data class would be trained on separate GPUs. CIFAR10 dataset, the number of  processes spans the range from 1 to 10, whereas the number of processes, for the CIFAR100 dataset was set to 10, 20, 40, 80 and 100, and for the ImageNet1k dataset was set to 100, 200, 400, 800 and 1,000. The results for the scalability tests on CIFAR10 and CIFAR100 datasets are shown in Figure \ref{scalability_pic} { by reporting the runtime of the slowest processor}. The trend of the wall-clock time shows that the computational time to complete the training is not affected by the increasing number of data classes, as long as the computational workload for each process stays fixed, thus showing that weak scaling is obtained by the training of parallel W-CGANs on all three datasets. 
On the ImageNet1k dataset, we notice a deterioration of the weak scaling performance of the parallel W-CGANs for 200 processes and beyond. 
Since the ImageNet1k datasets has balanced classes, the same workload is assigned to each W-CGANs. Therefore, the deteriorated scalability cannot be attributed to a non-uniform work load balance of the algorithm. Additional studies are being currently conducted to understand what interconnectivity properties in the hardware can cause the deterioration of the scaling. One simple workaround is at the job scheduler level by running multiple jobs (all below 200 GPUs), so that the overall time-to-solution still abides by ideal weak scaling performance.   
The average GPU utilization is 93.8\% with a standard deviation of 0.7\%, and the memory utilization is 7,839 MiB.

\section{Conclusions and future developments}
We used the Wasserstein metric to stabilize the {parallel} training of CGANs, thereby maintaining scalability while avoiding both cycling and mode collapse. Numerical results have been presented using the three open-source benchmark image datasets CIFAR10, CIFAR100, and ImageNet1k. The fact that CIFAR100 has fewer image data per class makes the GANs training more difficult than for CIFAR10, but the use of the Wasserstein metric stabilizes the {parallel} training of W-CGANs, which leads to better results as it is validated by higher IS score, lower FID score, and better quality images than what previously obtained with a {parallel} training of DC-CGANs \cite{LupoPasini2021}. Weak scaling plots reporting the wall-clock time of the slowest process show that the stabilization obtained with W-CGANs does not compromise the scalability of the approach on CIFAR10 and CIFAR100 datasets. For ImageNet1k scalability performance start deteriorating for more than 200 concurrent processes.  

{Future} work will extend this study to more complex but also more accurate neural network architectures, such as auxiliary classifier GANs (AC-GANs) \cite{pmlr-v70-odena17a}, residual neural networks (ResNet) \cite{Wang2017GenerativeAN}, and self-attention generative adversarial neural networks (SAGANs) \cite{pmlr-v97-zhang19d}, {We will also introduce regular communications between the model replicas to allow distributed training in situations where interpolating between data from different classes is useful}.



\section*{Acknowledgements}
Massimiliano Lupo Pasini thanks Dr. Vladimir Protopopescu for his valuable feedback in the preparation of this manuscript.
This work was supported in part by the Office of Science of the Department of Energy and by the Laboratory Directed Research and Development (LDRD) Program of Oak Ridge National Laboratory. 
This research is sponsored by the Artificial Intelligence Initiative as part of the Laboratory Directed Research and Development Program of Oak Ridge National Laboratory, managed by UT-Battelle, LLC, for the US Department of Energy under contract DE-AC05-00OR22725.
This work used resources of the Oak Ridge Leadership Computing Facility, which is supported by the Office of Science of the U.S. Department of Energy under Contract No. DE-AC05-00OR22725.

%
\section*{Declarations}
The authors declare that they have no conflict of interest.
\\[3mm]
Data sharing not applicable to this article as no datasets were generated or analysed during the current study.

\bibliographystyle{spbasic}      
\bibliography{sn-bibliography}


\begin{thebibliography}{31}
\ifx \bisbn   \undefined \def \bisbn  #1{ISBN #1}\fi
\ifx \binits  \undefined \def \binits#1{#1}\fi
\ifx \bauthor  \undefined \def \bauthor#1{#1}\fi
\ifx \batitle  \undefined \def \batitle#1{#1}\fi
\ifx \bjtitle  \undefined \def \bjtitle#1{#1}\fi
\ifx \bvolume  \undefined \def \bvolume#1{\textbf{#1}}\fi
\ifx \byear  \undefined \def \byear#1{#1}\fi
\ifx \bissue  \undefined \def \bissue#1{#1}\fi
\ifx \bfpage  \undefined \def \bfpage#1{#1}\fi
\ifx \blpage  \undefined \def \blpage #1{#1}\fi
\ifx \burl  \undefined \def \burl#1{\textsf{#1}}\fi
\ifx \doiurl  \undefined \def \doiurl#1{\url{https://doi.org/#1}}\fi
\ifx \betal  \undefined \def \betal{\textit{et al.}}\fi
\ifx \binstitute  \undefined \def \binstitute#1{#1}\fi
\ifx \binstitutionaled  \undefined \def \binstitutionaled#1{#1}\fi
\ifx \bctitle  \undefined \def \bctitle#1{#1}\fi
\ifx \beditor  \undefined \def \beditor#1{#1}\fi
\ifx \bpublisher  \undefined \def \bpublisher#1{#1}\fi
\ifx \bbtitle  \undefined \def \bbtitle#1{#1}\fi
\ifx \bedition  \undefined \def \bedition#1{#1}\fi
\ifx \bseriesno  \undefined \def \bseriesno#1{#1}\fi
\ifx \blocation  \undefined \def \blocation#1{#1}\fi
\ifx \bsertitle  \undefined \def \bsertitle#1{#1}\fi
\ifx \bsnm \undefined \def \bsnm#1{#1}\fi
\ifx \bsuffix \undefined \def \bsuffix#1{#1}\fi
\ifx \bparticle \undefined \def \bparticle#1{#1}\fi
\ifx \barticle \undefined \def \barticle#1{#1}\fi
\bibcommenthead
\ifx \bconfdate \undefined \def \bconfdate #1{#1}\fi
\ifx \botherref \undefined \def \botherref #1{#1}\fi
\ifx \url \undefined \def \url#1{\textsf{#1}}\fi
\ifx \bchapter \undefined \def \bchapter#1{#1}\fi
\ifx \bbook \undefined \def \bbook#1{#1}\fi
\ifx \bcomment \undefined \def \bcomment#1{#1}\fi
\ifx \oauthor \undefined \def \oauthor#1{#1}\fi
\ifx \citeauthoryear \undefined \def \citeauthoryear#1{#1}\fi
\ifx \endbibitem  \undefined \def \endbibitem {}\fi
\ifx \bconflocation  \undefined \def \bconflocation#1{#1}\fi
\ifx \arxivurl  \undefined \def \arxivurl#1{\textsf{#1}}\fi
\csname PreBibitemsHook\endcsname

\bibitem{goodfellow_generative_2014}
\begin{bchapter}
\bauthor{\bsnm{Goodfellow}, \binits{I.}},
\bauthor{\bsnm{Pouget-Abadie}, \binits{J.}},
\bauthor{\bsnm{Mirza}, \binits{M.}},
\bauthor{\bsnm{Xu}, \binits{B.}},
\bauthor{\bsnm{Warde-Farley}, \binits{D.}},
\bauthor{\bsnm{Ozair}, \binits{S.}},
\bauthor{\bsnm{Courville}, \binits{A.}},
\bauthor{\bsnm{Bengio}, \binits{Y.}}:
\bctitle{Generative adversarial nets}.
In: \beditor{\bsnm{Ghahramani}, \binits{Z.}},
\beditor{\bsnm{Welling}, \binits{M.}},
\beditor{\bsnm{Cortes}, \binits{C.}},
\beditor{\bsnm{Lawrence}, \binits{N.}},
\beditor{\bsnm{Weinberger}, \binits{K.Q.}} (eds.)
\bbtitle{Advances in Neural Information Processing Systems},
vol. \bseriesno{27}.
\bpublisher{Curran Associates, Inc.},
\blocation{Palais des Congrès de Montréal, Montréal, Canada}
(\byear{2014}).
\burl{https://proceedings.neurips.cc/paper/2014/file/5ca3e9b122f61f8f06494c97b1afccf3-Paper.pdf}
\end{bchapter}
\endbibitem

\bibitem{radford_unsupervised_2016}
\begin{bchapter}
\bauthor{\bsnm{Radford}, \binits{A.}},
\bauthor{\bsnm{Metz}, \binits{L.}},
\bauthor{\bsnm{Chintala}, \binits{S.}}:
\bctitle{Unsupervised representation learning with deep convolutional
  generative adversarial networks}.
In: \beditor{\bsnm{Bengio}, \binits{Y.}},
\beditor{\bsnm{LeCun}, \binits{Y.}} (eds.)
\bbtitle{4th International Conference on Learning Representations, {ICLR} 2016,
  Conference Track Proceedings},
\bconflocation{San Juan, Puerto Rico, May 2-4, 2016}
(\byear{2016}).
\burl{http://arxiv.org/abs/1511.06434}
\end{bchapter}
\endbibitem

\bibitem{salimans_improved_2016}
\begin{botherref}
\oauthor{\bsnm{Salimans}, \binits{T.}},
\oauthor{\bsnm{Goodfellow}, \binits{I.}},
\oauthor{\bsnm{Zaremba}, \binits{W.}},
\oauthor{\bsnm{Cheung}, \binits{V.}},
\oauthor{\bsnm{Radford}, \binits{A.}},
\oauthor{\bsnm{Chen}, \binits{X.}}:
Improved techniques for training GANs.
arXiv
(2016).
\doiurl{10.48550/ARXIV.1606.03498}.
\url{https://arxiv.org/abs/1606.03498}
\end{botherref}
\endbibitem

\bibitem{bertsekas_multiagent_2019}
\begin{botherref}
\oauthor{\bsnm{Bertsekas}, \binits{D.}}:
Multiagent rollout algorithms and reinforcement learning.
arXiv
(2019).
\doiurl{10.48550/ARXIV.1910.00120}.
\url{https://arxiv.org/abs/1910.00120}
\end{botherref}
\endbibitem

\bibitem{mertikopoulos_2017}
\begin{botherref}
\oauthor{\bsnm{Mertikopoulos}, \binits{P.}},
\oauthor{\bsnm{Papadimitriou}, \binits{C.}},
\oauthor{\bsnm{Piliouras}, \binits{G.}}:
Cycles in adversarial regularized learning.
In: Proceedings of the 2018 Annual ACM-SIAM Symposium on Discrete Algorithms
  (SODA),
pp. 2703--2717.
\doiurl{10.1137/1.9781611975031.172}.
\url{https://epubs.siam.org/doi/abs/10.1137/1.9781611975031.172}
\end{botherref}
\endbibitem

\bibitem{NIPS2017_165a59f7}
\begin{bchapter}
\bauthor{\bsnm{Hazan}, \binits{E.}},
\bauthor{\bsnm{Singh}, \binits{K.}},
\bauthor{\bsnm{Zhang}, \binits{C.}}:
\bctitle{Learning linear dynamical systems via spectral filtering}.
In: \beditor{\bsnm{Guyon}, \binits{I.}},
\beditor{\bsnm{Luxburg}, \binits{U.V.}},
\beditor{\bsnm{Bengio}, \binits{S.}},
\beditor{\bsnm{Wallach}, \binits{H.}},
\beditor{\bsnm{Fergus}, \binits{R.}},
\beditor{\bsnm{Vishwanathan}, \binits{S.}},
\beditor{\bsnm{Garnett}, \binits{R.}} (eds.)
\bbtitle{Advances in Neural Information Processing Systems},
vol. \bseriesno{30}.
\bpublisher{Curran Associates, Inc.},
\blocation{Long Beach Convention \& Entertainment Center, Long Beach,
  California, USA}
(\byear{2017}).
\burl{https://proceedings.neurips.cc/paper/2017/file/165a59f7cf3b5c4396ba65953d679f17-Paper.pdf}
\end{bchapter}
\endbibitem

\bibitem{LupoPasini2021}
\begin{botherref}
\oauthor{\bsnm{Lupo~Pasini}, \binits{M.}},
\oauthor{\bsnm{Gabbi}, \binits{V.}},
\oauthor{\bsnm{Yin}, \binits{J.}},
\oauthor{\bsnm{Perotto}, \binits{S.}},
\oauthor{\bsnm{Laanait}, \binits{N.}}:
Scalable balanced training of conditional generative adversarial neural
  networks on image data.
Journal of Supercomputing
\textbf{77}(11)
(2021).
\doiurl{10.1007/s11227-021-03808-2}
\end{botherref}
\endbibitem

\bibitem{mirza_conditional_2014}
\begin{botherref}
\oauthor{\bsnm{Mirza}, \binits{M.}},
\oauthor{\bsnm{Osindero}, \binits{S.}}:
Conditional Generative Adversarial Nets.
arXiv
(2014).
\doiurl{10.48550/ARXIV.1411.1784}.
\url{https://arxiv.org/abs/1411.1784}
\end{botherref}
\endbibitem

\bibitem{yang2018diversitysensitive}
\begin{bchapter}
\bauthor{\bsnm{Yang}, \binits{D.}},
\bauthor{\bsnm{Hong}, \binits{S.}},
\bauthor{\bsnm{Jang}, \binits{Y.}},
\bauthor{\bsnm{Zhao}, \binits{T.}},
\bauthor{\bsnm{Lee}, \binits{H.}}:
\bctitle{Diversity-sensitive conditional generative adversarial networks}.
In: \bbtitle{International Conference on Learning Representations}.
\bpublisher{OpenReview.net},
\blocation{New Orleans, LA, USA}
(\byear{2019})
\end{bchapter}
\endbibitem

\bibitem{Zhou2020}
\begin{botherref}
\oauthor{\bsnm{Zhou}, \binits{P.}},
\oauthor{\bsnm{Xie}, \binits{L.}},
\oauthor{\bsnm{Zhang}, \binits{X.}},
\oauthor{\bsnm{Ni}, \binits{B.}},
\oauthor{\bsnm{Tian}, \binits{Q.}}:
Searching towards Class-Aware Generators for Conditional Generative Adversarial
  Networks.
arXiv
(2020).
\doiurl{10.48550/ARXIV.2006.14208}.
\url{https://arxiv.org/abs/2006.14208}
\end{botherref}
\endbibitem

\bibitem{R1}
\begin{barticle}
\bauthor{\bsnm{Zhang}, \binits{H.}},
\bauthor{\bsnm{Sindagi}, \binits{V.}},
\bauthor{\bsnm{Pate}, \binits{V.M.}}:
\batitle{Image de-raining using a conditional generative adversarial network}.
\bjtitle{IEEE Transactions on Circuits and Systems for Video Technology}
(\byear{2020}).
\doiurl{10.1109/TCSVT.2019.2920407}.
\bcomment{10.1109/TCSVT.2019.2920407}
\end{barticle}
\endbibitem

\bibitem{miyato2018cgans}
\begin{bchapter}
\bauthor{\bsnm{Miyato}, \binits{T.}},
\bauthor{\bsnm{Koyama}, \binits{M.}}:
\bctitle{{cGANs} with projection discriminator}.
In: \bbtitle{6th International Conference on Learning Representations, {ICLR}
  2018, Vancouver, BC, Canada, April 30 - May 3, 2018, Conference Track
  Proceedings}.
\bpublisher{OpenReview.net},
\blocation{Vancouver Convention Center, Vancouver, Canada}
(\byear{2018}).
\burl{https://openreview.net/forum?id=ByS1VpgRZ}
\end{bchapter}
\endbibitem

\bibitem{DBLP:journals/corr/abs-1912-04216}
\begin{botherref}
\oauthor{\bsnm{Kavalerov}, \binits{I.}},
\oauthor{\bsnm{Czaja}, \binits{W.}},
\oauthor{\bsnm{Chellappa}, \binits{R.}}:
cGANs with Multi-Hinge Loss.
arXiv
(2019).
\doiurl{10.48550/ARXIV.1912.04216}.
\url{https://arxiv.org/abs/1912.04216}
\end{botherref}
\endbibitem

\bibitem{Heusel2020}
\begin{bchapter}
\bauthor{\bsnm{Heusel}, \binits{M.}},
\bauthor{\bsnm{Ramsauer}, \binits{H.}},
\bauthor{\bsnm{Unterthiner}, \binits{T.}},
\bauthor{\bsnm{Nessler}, \binits{B.}},
\bauthor{\bsnm{Hochreiter}, \binits{S.}}:
\bctitle{{GANs} trained by a two time-scale update rule converge to a local
  nash equilibrium}.
In: \bbtitle{Proceedings of the 31st International Conference on Neural
  Information Processing Systems}.
\bsertitle{NIPS'17},
pp. \bfpage{6629}--\blpage{6640}.
\bpublisher{Curran Associates Inc.},
\blocation{Red Hook, NY, USA}
(\byear{2017})
\end{bchapter}
\endbibitem

\bibitem{cifar10}
\begin{botherref}
\oauthor{\bsnm{Krizhevsky}, \binits{A.}},
\oauthor{\bsnm{Nair}, \binits{V.}},
\oauthor{\bsnm{Hinton}, \binits{G.}}:
Cifar-10 {(Canadian Institute for Advanced Research)}
(2009)
\end{botherref}
\endbibitem

\bibitem{cifar100}
\begin{botherref}
\oauthor{\bsnm{Krizhevsky}, \binits{A.}},
\oauthor{\bsnm{Nair}, \binits{V.}},
\oauthor{\bsnm{Hinton}, \binits{G.}}:
Cifar-100 {(Canadian Institute for Advanced Research)}
(2009)
\end{botherref}
\endbibitem

\bibitem{imagenet}
\begin{botherref}
{ImageNet}.
\texttt{http://image-net.org/}
\end{botherref}
\endbibitem

\bibitem{KL}
\begin{barticle}
\bauthor{\bsnm{Kullback}, \binits{S.}},
\bauthor{\bsnm{Leibler}, \binits{R.A.}}:
\batitle{On information and sufficiency}.
\bjtitle{Annals of Mathematical Statistics}
\bvolume{22}(\bissue{1}),
\bfpage{79}--\blpage{86}
(\byear{1951})
\end{barticle}
\endbibitem

\bibitem{Lin1991}
\begin{barticle}
\bauthor{\bsnm{Lin}, \binits{J.}}:
\batitle{Divergence measures based on the {Shannon} entropy}.
\bjtitle{IEEE Transactions on Information Theory}
\bvolume{37}(\bissue{1}),
\bfpage{145}--\blpage{151}
(\byear{1991}).
\doiurl{10.1109/18.61115}
\end{barticle}
\endbibitem

\bibitem{kl_wasserstein_comparison}
\begin{bchapter}
\bauthor{\bsnm{Belavkin}, \binits{R.V.}}:
\bctitle{Relation between the {Kantorovich--Wasserstein} metric and the
  {Kullback--Leibler} divergence}.
In: \beditor{\bsnm{Ay}, \binits{N.}},
\beditor{\bsnm{Gibilisco}, \binits{P.}},
\beditor{\bsnm{Mat{\'u}{\v{s}}}, \binits{F.}} (eds.)
\bbtitle{Information Geometry and Its Applications},
pp. \bfpage{363}--\blpage{373}.
\bpublisher{Springer},
\blocation{Cham}
(\byear{2018})
\end{bchapter}
\endbibitem

\bibitem{Lipschitz}
\begin{bchapter}
\bauthor{\bsnm{Scaman}, \binits{K.}},
\bauthor{\bsnm{Virmaux}, \binits{A.}}:
\bctitle{Lipschitz regularity of deep neural networks: analysis and efficient
  estimation}.
In: \bbtitle{Proceedings of the 32nd International Conference on Neural
  Information Processing Systems}.
\bsertitle{NIPS'18},
pp. \bfpage{3839}--\blpage{3848}.
\bpublisher{Curran Associates Inc.},
\blocation{Red Hook, NY, USA}
(\byear{2018})
\end{bchapter}
\endbibitem

\bibitem{brock2018large}
\begin{bchapter}
\bauthor{\bsnm{Brock}, \binits{A.}},
\bauthor{\bsnm{Donahue}, \binits{J.}},
\bauthor{\bsnm{Simonyan}, \binits{K.}}:
\bctitle{Large scale {GAN} training for high fidelity natural image synthesis}.
In: \bbtitle{International Conference on Learning Representations}.
\bpublisher{OpenReview.net},
\blocation{New Orleans, LA, USA}
(\byear{2019}).
\burl{https://openreview.net/forum?id=B1xsqj09Fm}
\end{bchapter}
\endbibitem

\bibitem{refId0}
\begin{barticle}
\bauthor{\bsnm{{Vlimant, Jean-Roch}}},
\bauthor{\bsnm{{Pantaleo, Felice}}},
\bauthor{\bsnm{{Pierini, Maurizio}}},
\bauthor{\bsnm{{Loncar, Vladimir}}},
\bauthor{\bsnm{{Vallecorsa, Sofia}}},
\bauthor{\bsnm{{Anderson, Dustin}}},
\bauthor{\bsnm{{Nguyen, Thong}}},
\bauthor{\bsnm{{Zlokapa, Alexander}}}:
\batitle{Large-scale distributed training applied to generative adversarial
  networks for calorimeter simulation}.
\bjtitle{EPJ Web of Conferences}
\bvolume{214},
\bfpage{06025}
(\byear{2019}).
\doiurl{10.1051/epjconf/201921406025}
\end{barticle}
\endbibitem

\bibitem{NIPS2012_6aca9700}
\begin{bchapter}
\bauthor{\bsnm{Dean}, \binits{J.}},
\bauthor{\bsnm{Corrado}, \binits{G.}},
\bauthor{\bsnm{Monga}, \binits{R.}},
\bauthor{\bsnm{Chen}, \binits{K.}},
\bauthor{\bsnm{Devin}, \binits{M.}},
\bauthor{\bsnm{Mao}, \binits{M.}},
\bauthor{\bsnm{Ranzato}, \binits{M.a.}},
\bauthor{\bsnm{Senior}, \binits{A.}},
\bauthor{\bsnm{Tucker}, \binits{P.}},
\bauthor{\bsnm{Yang}, \binits{K.}},
\bauthor{\bsnm{Le}, \binits{Q.}},
\bauthor{\bsnm{Ng}, \binits{A.}}:
\bctitle{Large scale distributed deep networks}.
In: \beditor{\bsnm{Pereira}, \binits{F.}},
\beditor{\bsnm{Burges}, \binits{C.J.}},
\beditor{\bsnm{Bottou}, \binits{L.}},
\beditor{\bsnm{Weinberger}, \binits{K.Q.}} (eds.)
\bbtitle{Advances in Neural Information Processing Systems},
vol. \bseriesno{25}.
\bpublisher{Curran Associates, Inc.},
\blocation{Lake Tahoe, NV, USA}
(\byear{2012}).
\burl{https://proceedings.neurips.cc/paper/2012/file/6aca97005c68f1206823815f66102863-Paper.pdf}
\end{bchapter}
\endbibitem

\bibitem{DBLP:conf/nips/Liu0MCRYD20}
\begin{bchapter}
\bauthor{\bsnm{Liu}, \binits{M.}},
\bauthor{\bsnm{Zhang}, \binits{W.}},
\bauthor{\bsnm{Mroueh}, \binits{Y.}},
\bauthor{\bsnm{Cui}, \binits{X.}},
\bauthor{\bsnm{Ross}, \binits{J.}},
\bauthor{\bsnm{Yang}, \binits{T.}},
\bauthor{\bsnm{Das}, \binits{P.}}:
\bctitle{A decentralized parallel algorithm for training generative adversarial
  nets}.
In: \beditor{\bsnm{Larochelle}, \binits{H.}},
\beditor{\bsnm{Ranzato}, \binits{M.}},
\beditor{\bsnm{Hadsell}, \binits{R.}},
\beditor{\bsnm{Balcan}, \binits{M.}},
\beditor{\bsnm{Lin}, \binits{H.}} (eds.)
\bbtitle{Advances in Neural Information Processing Systems 33: Annual
  Conference on Neural Information Processing Systems 2020, NeurIPS 2020,
  December 6-12, 2020, Virtual}
(\byear{2020}).
\burl{https://proceedings.neurips.cc/paper/2020/hash/7e0a0209b929d097bd3e8ef30567a5c1-Abstract.html}
\end{bchapter}
\endbibitem

\bibitem{barratt_note_2018}
\begin{botherref}
\oauthor{\bsnm{Barratt}, \binits{S.}},
\oauthor{\bsnm{Sharma}, \binits{R.}}:
A Note on the Inception Score.
arXiv
(2018).
\doiurl{10.48550/ARXIV.1801.01973}.
\url{https://arxiv.org/abs/1801.01973}
\end{botherref}
\endbibitem

\bibitem{summit}
\begin{botherref}
\oauthor{\bsnm{{Oak Ridge Leadership Facility}}}:
Summit - Oak Ridge National Laboratory's 200 petaflop supercomputer.
\url{https://www.olcf.ornl.gov/olcf-resources/compute-systems/summit/}.
Accessed: 2021-08-21
(2018)
\end{botherref}
\endbibitem

\bibitem{paszke_pytorch_2019}
\begin{bchapter}
\bauthor{\bsnm{Paszke}, \binits{A.}},
\bauthor{\bsnm{Gross}, \binits{S.}},
\bauthor{\bsnm{Massa}, \binits{F.}},
\bauthor{\bsnm{Lerer}, \binits{A.}},
\bauthor{\bsnm{Bradbury}, \binits{J.}},
\bauthor{\bsnm{Chanan}, \binits{G.}},
\bauthor{\bsnm{Killeen}, \binits{T.}},
\bauthor{\bsnm{Lin}, \binits{Z.}},
\bauthor{\bsnm{Gimelshein}, \binits{N.}},
\bauthor{\bsnm{Antiga}, \binits{L.}},
\bauthor{\bsnm{Desmaison}, \binits{A.}},
\bauthor{\bsnm{Kopf}, \binits{A.}},
\bauthor{\bsnm{Yang}, \binits{E.}},
\bauthor{\bsnm{DeVito}, \binits{Z.}},
\bauthor{\bsnm{Raison}, \binits{M.}},
\bauthor{\bsnm{Tejani}, \binits{A.}},
\bauthor{\bsnm{Chilamkurthy}, \binits{S.}},
\bauthor{\bsnm{Steiner}, \binits{B.}},
\bauthor{\bsnm{Fang}, \binits{L.}},
\bauthor{\bsnm{Bai}, \binits{J.}},
\bauthor{\bsnm{Chintala}, \binits{S.}}:
\bctitle{{PyTorch}: {An} {Imperative} {Style}, {High}-{Performance} {Deep}
  {Learning} {Library}}.
In: \beditor{\bsnm{Wallach}, \binits{H.}},
\beditor{\bsnm{Larochelle}, \binits{H.}},
\beditor{\bsnm{Beygelzimer}, \binits{A.}},
\beditor{\bsnm{Alché-Buc}, \binits{F.}},
\beditor{\bsnm{Fox}, \binits{E.}},
\beditor{\bsnm{Garnett}, \binits{R.}} (eds.)
\bbtitle{Proceedings of the 33rd International Conference on Neural Information
  Processing Systems},
pp. \bfpage{8026}--\blpage{8037}.
\bpublisher{Curran Associates, Inc.},
\blocation{Red Hook, NY, USA}
(\byear{2019})
\end{bchapter}
\endbibitem

\bibitem{pmlr-v70-odena17a}
\begin{bchapter}
\bauthor{\bsnm{Odena}, \binits{A.}},
\bauthor{\bsnm{Olah}, \binits{C.}},
\bauthor{\bsnm{Shlens}, \binits{J.}}:
\bctitle{{C}onditional {I}mage {S}ynthesis with {A}uxiliary {C}lassifier
  {GAN}s}.
In: \beditor{\bsnm{Precup}, \binits{D.}},
\beditor{\bsnm{Teh}, \binits{Y.W.}} (eds.)
\bbtitle{Proceedings of the 34th International Conference on Machine Learning}.
\bsertitle{Proceedings of Machine Learning Research},
vol. \bseriesno{70},
pp. \bfpage{2642}--\blpage{2651}.
\bpublisher{PMLR},
\blocation{International Convention Centre, Sydney, Australia}
(\byear{2017}).
\burl{http://proceedings.mlr.press/v70/odena17a.html}
\end{bchapter}
\endbibitem

\bibitem{Wang2017GenerativeAN}
\begin{botherref}
\oauthor{\bsnm{Wang}, \binits{M.}},
\oauthor{\bsnm{Li}, \binits{H.}},
\oauthor{\bsnm{Li}, \binits{F.}}:
Generative adversarial network based on Resnet for conditional image
  restoration.
arXiv
(2017).
\doiurl{10.48550/ARXIV.1707.04881}.
\url{https://arxiv.org/abs/1707.04881}
\end{botherref}
\endbibitem

\bibitem{pmlr-v97-zhang19d}
\begin{bchapter}
\bauthor{\bsnm{Zhang}, \binits{H.}},
\bauthor{\bsnm{Goodfellow}, \binits{I.}},
\bauthor{\bsnm{Metaxas}, \binits{D.}},
\bauthor{\bsnm{Odena}, \binits{A.}}:
\bctitle{{S}elf-attention generative adversarial networks}.
In: \beditor{\bsnm{Chaudhuri}, \binits{K.}},
\beditor{\bsnm{Salakhutdinov}, \binits{R.}} (eds.)
\bbtitle{Proceedings of the 36th International Conference on Machine Learning}.
\bsertitle{Proceedings of Machine Learning Research},
vol. \bseriesno{97},
pp. \bfpage{7354}--\blpage{7363}.
\bpublisher{PMLR},
\blocation{Long Beach Convention \& Entertainment Center, Long Beach,
  California, USA}
(\byear{2019})
\end{bchapter}
\endbibitem

\end{thebibliography}

\end{document}